\documentclass[10pt, conference, compsocconf]{IEEEtran}
\usepackage{cite}

\ifCLASSINFOpdf
  \usepackage[pdftex]{graphicx}
  \graphicspath{{pdf_figures/}}
  \DeclareGraphicsExtensions{.pdf}
\else
\fi
\usepackage[cmex10]{amsmath}
\usepackage{array}

\usepackage{mdwmath}
\usepackage{mdwtab}

\usepackage{eqparbox}

\usepackage[caption=false,font=footnotesize]{subfig}
\usepackage{url}

\usepackage{epsfig}
\usepackage{amssymb}
\usepackage[utf8]{inputenc}
\usepackage[english]{babel}
\usepackage [autostyle, english = american]{csquotes}
\MakeOuterQuote{"}

\usepackage{algorithm,tabularx}
\usepackage{algpseudocode}

\usepackage{multirow}
\usepackage{makecell}

\usepackage{xspace}

\algrenewcommand\algorithmicrequire{\textbf{Input:}}
\algrenewcommand\algorithmicensure{\textbf{Output:}}

\makeatletter
\newcommand{\multiline}[1]{%
  \begin{tabularx}{\dimexpr\linewidth-\ALG@thistlm}[t]{@{}X@{}}
    #1
  \end{tabularx}
}
\makeatother

\newcounter{algsubstate}

\makeatletter
\algnewcommand{\ProperState}[1]{\Statex \hskip\ALG@thistlm #1}
\makeatother
\newenvironment{algsubstates}
  {\setcounter{algsubstate}{0}%
   \renewcommand{\State}{%
     \refstepcounter{algsubstate}%
     \ProperState{\small{\alph{algsubstate}}:}\space}}
  {}

\DeclareMathOperator{\EX}{\mathbb{E}} 

\newcommand{\RMSprop}{\mathrm{RMSprop}}

\newcommand\trajfigheight{0.15}
\newcommand\trajfigwidth{0.24}

\makeatletter
\DeclareRobustCommand\onedot{\futurelet\@let@token\@onedot}
\def\@onedot{\ifx\@let@token.\else.\null\fi\xspace}

\def\etal{\emph{et al}\onedot}
\makeatother

\begin{document}
%
\title{Adversarially Learned Abnormal Trajectory Classifier}


\author{\IEEEauthorblockN{Pankaj Raj Roy, Guillaume-Alexandre Bilodeau}
\IEEEauthorblockA{LITIV lab.\\
Polytechnique Montr\'{e}al,
Montr\'{e}al, Canada\\
pankaj-raj.roy@polymtl.ca, gabilodeau@polymtl.ca}
}


%


\maketitle

\begin{abstract}
We address the problem of abnormal event detection from trajectory data. In this paper, a new adversarial approach is proposed for building a deep neural network binary classifier, trained in an unsupervised fashion, that can distinguish normal from abnormal trajectory-based events without the need for setting manual detection threshold. Inspired by the generative adversarial network (GAN) framework, our GAN version is a discriminative one in which the discriminator is trained to distinguish normal and abnormal trajectory reconstruction errors given by a deep autoencoder. With urban traffic videos and their associated trajectories, our proposed method gives the best accuracy for abnormal trajectory detection. In addition, our model can easily be generalized for abnormal trajectory-based event detection and can still yield the best behavioural detection results as demonstrated on the CAVIAR dataset.
\end{abstract}


%
\IEEEpeerreviewmaketitle

\section{Introduction}
Nowadays, the collection of user data is increasing exponentially. With this huge amount of data, many end users struggle to find the most efficient way of interpreting it. One of the most challenging tasks is to learn and detect unusual information patterns from the observed data. This kind of information can be thought as any form of observations that do not follow the usual ones and that can also look suspicious. A popular application of anomaly detection is the detection of abnormal events in video surveillance  \cite{ref_article2,ref_article5,ref_article8,ref_article6} in which the main purpose is to identify all the pixel groups that deviate from the ordinarily observed groups. The main issue behind it is the fact that there are many possible abnormal events and detecting most of them requires a large amount of normal and abnormal training data.

\begin{figure}[t]
    \centering
    \includegraphics[width=1\linewidth]{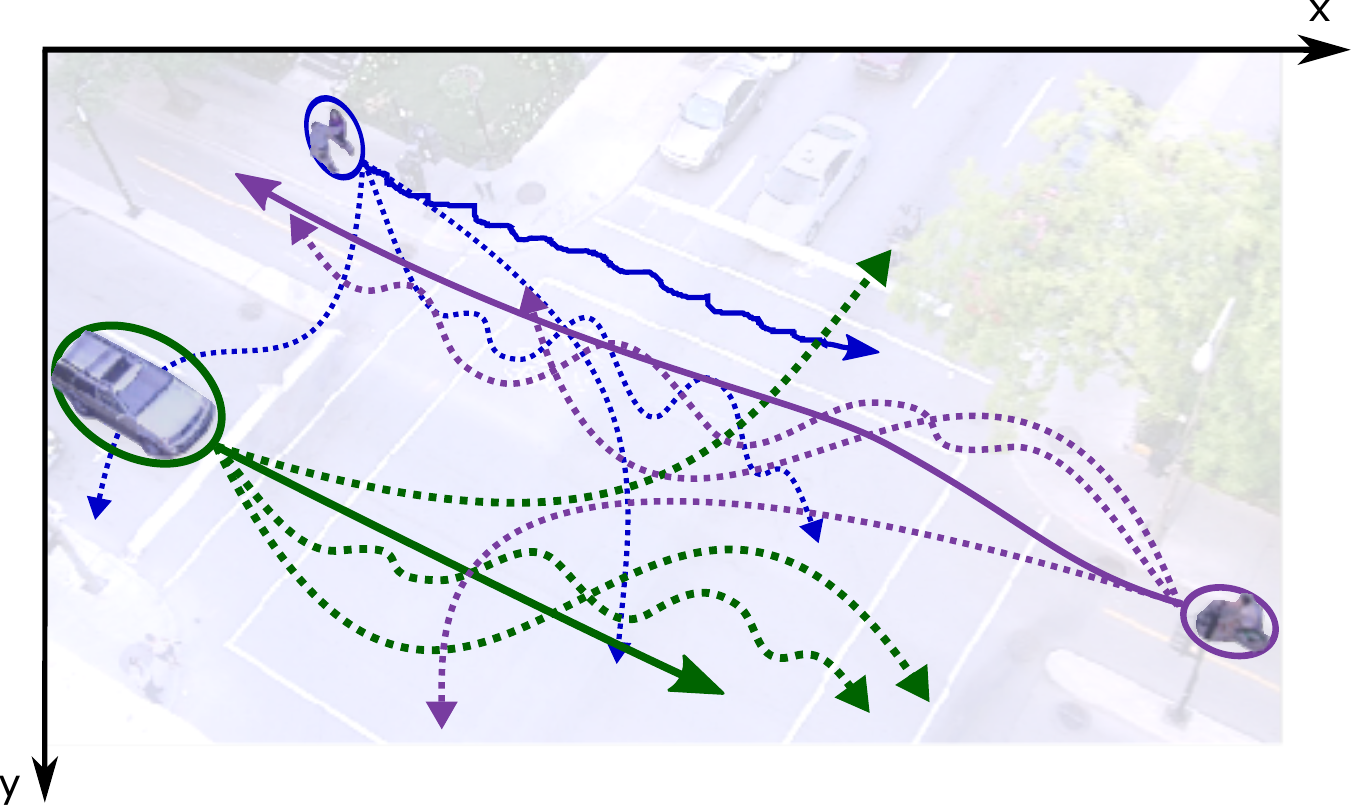}
    \caption{Example of normal (solid lines) and abnormal (dotted lines) trajectories of the road users including pedestrians (blue), cars (green), and bikes (purple), where the directions are illustrated by coloured arrows.}
    \label{fig:traj_example}
\end{figure}

In the case of abnormal road user trajectories as shown in figure \ref{fig:traj_example}, a significant amount of frames is required in order to form a trajectory and, by doing so, the pre-processing step of transforming multiple object locations into their corresponding trajectories significantly lessens the amount of abnormal training data required for any kinds of statistical or machine learning approaches. A solution would be to learn usual representation and detect abnormalities as outliers \cite{ref_article2,ref_article5,ref_article8,ref_article6,article1}. This, however, needs the computation of a threshold value that separates normal from abnormal data using a generally defined formula. The latter can be context dependent and, therefore, the formula itself might need to be manually tuned in order to have a better threshold value.

In this paper, we propose the idea of using an adversarial network which basically transforms a one-class deep autoencoder (DAE), like the one used in \cite{ref_deep_autoencoder}, which learns solely from normal data, into a two-class network that can classify normal and abnormal trajectories without the need of setting manually a detection threshold. We use a similar data structure and deep autoencoder model to that of Roy and Bilodeau~\cite{article1} proposed, but, instead of computing the threshold value that separates normal from abnormal data, we integrate the pretrained DAE into a deep fully-connected generative adversarial network (GAN). The GAN is designed in such a way to automatically find the best classification boundary between normal and abnormal data by generating realistic abnormal data for training the discriminator. We applied our method for detecting abnormal road user trajectories, as well as for detecting suspicious behaviours in video surveillance. In both cases, our proposed approach outperforms previous work.

Our contributions are: 1) a GAN model in which the generator produces some mean square trajectory reconstruction errors in order for the discriminator to better distinguish abnormal trajectories, 2) an algorithm for training efficiently our GAN version which enables the discriminative model to learn the most realistic abnormal data while ensuring that normal data is correctly classified, and 3) a demonstration that, by using the same dataset as \cite{article1}, our proposed method gives better results for the detection of abnormal road user trajectories and that our model can be generalized for other abnormal event detection applications.

\section{Related Work}
Considering the hypothesis that abnormal events are outliers, many authors recently suggested the idea of using an autoencoder that outputs reconstruction errors to distinguish normal and abnormal data. Ribeiro \etal~\cite{ref_article8} trained a convolutional autoencoder (CAE) model solely using normal training data for recognizing the structure of normal events. Therefore, the CAE network outputs the regularization of reconstruction errors (RRE) by taking as input the appearance features from a single frame extracted using a Canny edge detector and the motion features from optical flow. Normal and abnormal events are classified using a threshold value based on normal RRE. In \cite{ref_article_traditional_trajectory_method}, the authors use sparse reconstruction analysis by calculating the reconstruction residuals of the test trajectories by using L1-norm minimization and comparing those to the collected normal trajectory dictionary set constructed with the Least-squares Cubic Spline Curves Approximation (LCSCA) method. Again, a threshold is used for the classification decision. In the work of Roy and Bilodeau~\cite{article1}, a similar approach is used for classifying normal and abnormal road user trajectories using a threshold value computed through the mean squared error (MSE) of the reconstructed output of a DAE model trained with normal data only. Although the latter approach gives better precision than the most popular classical methods like a one-class support vector machine (OC-SVM) \cite{ref_Scholkopf}, it suffers from the following limitations: 1) the threshold does not take into account the structure of the variations of the squared reconstruction error (SRE) for each trajectory point, 2) stating that anything smaller or equal than the computed threshold value is considered as normal will not work with noisy input in the DAE model, and 3) the threshold has to be set manually for every new video.

To tackle the issue that comes with one-class models, some authors suggest incorporating GAN for generating plausible abnormalities and using the discriminative model for classifying normal and abnormal data. Sabokrou \etal~\cite{ref_AVID} proposed the idea of using a GAN where the first network acts as a generator which tries to fool the second network working as a regularity detector. Even though their method works well for finding irregularities in surveillance videos, it is difficult to make the generator converge in the case of trajectory data. Similarly, Sabokrou \etal~\cite{ref_Adversarially_Classifier_Sabokrou} uses the autoencoder (AE) model as the generator to adversarially learn reconstructed outputs that can fool the convolutional neural network (CNN) classifier. It also suffers from the same issue. Another work done by Gupta \etal~\cite{ref_social_GAN} suggests the social GAN approach where a recurrent discriminator is used for training a generator that outputs socially plausible trajectories. The generator is then used to predict future movements. Although this method can be applied for detecting abnormal trajectories, their proposition is only limited to pedestrians.

The GAN framework is a very interesting approach for generating data, especially when the latter is very limited. Several versions of the GAN have been proposed recently. Odena \etal~\cite{ref_AC-GAN} introduced Auxiliary Classifier GANs (AC-GANs) in order to improve the training of GANs for image synthesis, where the generated samples are based on the Gaussian noise and on the corresponding class label. Makhzani \etal~\cite{ref_Adversarial_Autoencoders} proposed the idea of Adversarial Autoencoder (AAE) in which the encoder of the framework learns to transform the data distribution into the prior one while respecting the formal autoencoder objective. Another one is the Bidirectional Generative Adversarial Network (BiGAN), suggested by Donahue \etal~\cite{ref_BiGAN}, in which the encoder and the generator of the framework learn to invert each other in order to fool the corresponding discriminator. Although these different varieties of GAN might be useful for image generation tasks, they appear to be very unstable during the training process in the case where we are applying them on a non-image data structure. In the case of the sparse data structure, where each element represents a corresponding point coordinate value, the data does not follow any particular distribution like image data. Therefore, it is difficult to generate realistic abnormal trajectory data with these methods.

\section{Proposed Approach}
Given the center of the bounding box of each object for each video frame, trajectory samples are extracted. The data is also augmented in order to increase the training efficiency of our deep neural network (DNN). These extracted and augmented trajectory samples, considered as normal, are then used in the training of a deep AE (DAE) network. We then train a GAN-based framework that adversarially learns an irregularity classifier solely by using the trained DAE and the same corresponding training samples.

\subsection{Problem Definition} \label{sec: 3.1}
We define trajectories as a collection of consecutive person/object positions and velocities. After the preprocessing steps that extract and augment the trajectory data, we use the following data structure for each sample \cite{article1}:

\begin{gather}
    \left[ u, x_1, y_1, v_{x_1}, v_{y_1}, x_2, y_2, v_{x_2}, v_{y_2}, ..., x_m, y_m, v_{x_m}, v_{y_m} \right] \textnormal{,} \label{eq:traj_struct}
\end{gather}

where $u$ is the object class label, $m$ is the number of trajectory points per sample, which are given in the 2D Cartesian coordinate system ($x$, $y$), $v_x$ and $v_y$ are the velocities of the corresponding object at a particular location according to $x$ and $y$ axes respectively. The problem to solve is to distinguish normal trajectory samples from abnormal ones. It is assumed that we observe only the normal trajectory during the training of our model.

\subsection{Background on Autoencoders and Adversarial Networks}
In our proposed method, two main machine learning frameworks are used for the detection of irregularities. Firstly, an autoencoder (AE) model, like the one used in \cite{ref_autoencoder}, is trained in order to reconstruct the input data. Therefore, any sample that deviates from the ones used during the training of the AE will result in a bad reconstruction, where the output does not resemble the original input data. Secondly, a generative adversarial network (GAN) \cite{Goodfellow:2014:GAN} that works, in part, as an abnormality generator that does not depend on any observed or handmade abnormal data, and in part, as a classifier which identifies if the provided input data is real or not.

\subsubsection{Autoencoder (AE)}
The general concept of an autoencoder is to be able to reconstruct the input data. The network does so by having two subnetworks stacked together. The original data first enters the encoder model in which the encoded data is generated. The encoded data is also known as the Compressed Feature vector (CFV) or the latent representation. After that, the CFV is passed through the decoder model in order to generate the reconstruction of the original input. The AE model learns to minimize the following objective function $\Theta_{\textnormal{AE}}$, shown in equation \ref{eq:obj_func_ae}, where $\psi(n)$ and $\hat{\psi}(n)$ are the input data and the corresponding reconstructed data.

\begin{gather}
    \Theta_{\textnormal{AE}} = \sum_{n=1}^N \left| \psi(n) - \hat{\psi}(n) \right|^2 \label{eq:obj_func_ae}
\end{gather}

It is worth mentioning that, although the objective function of an AE is the same as the Principal Component Analysis (PCA) method, the AE framework is more flexible and can handle nonlinear representations.

\subsubsection{Generative Adversarial Network (GAN)}
Inspired by a 2-player game-theoretic method, GANs are considered as the most efficient approach for generating realistic data. The original framework of GAN was first introduced by Goodfellow \etal~\cite{Goodfellow:2014:GAN} and the implementation was optimized for the image data type (like MNIST). The idea is to have a generator $G$ and a discriminator $D$ that are trained to fight each other, where $D$ classifies real/fake data knowing that $G$ generates fake one and where $G$ tries to fool $D$ by generating data that is indistinguishable from the real one. Globally, the networks of a GAN are trained via a two-player minimax game, represented by the optimization of the following objective function:

\begin{gather}
    \min_{G} \max_{D} \left( \resizebox{.6\hsize}{!}{$\EX\nolimits_{\psi \sim p_\psi} \left[ \log D(\psi) \right] + \EX\nolimits_{z \sim p_z} \left[1 - \log D(G(z)) \right]$} \right) \textnormal{,}
\end{gather}

where $p_\psi$ is the distribution of the real input data $\psi$ and $p_z$ describes the Gaussian distribution of the random noise $z$ used as an input of $G$. Basically, $D$ is learned by performing gradient ascent on that objective function and $G$ works by gradient descent on the same function. By following this alternating training approach, $G$ eventually fools $D$ with the generated realistic data $G(z)$.

Even though the training process of a GAN seems easily doable in theory, it can be very unstable when trying to make it work in practice, especially for trajectory data. Therefore, designing an algorithm for efficiently training a GAN model might be necessary for some tasks. In our case of abnormal trajectory detection, it was necessary to apply a training algorithm that makes the learning of the GAN model much more stable.

\subsection{Our Method: Adversarially Learned Reconstruction Error Classifier (ALREC)}
\begin{figure}[t]
    \centering
    \includegraphics[width=1\linewidth]{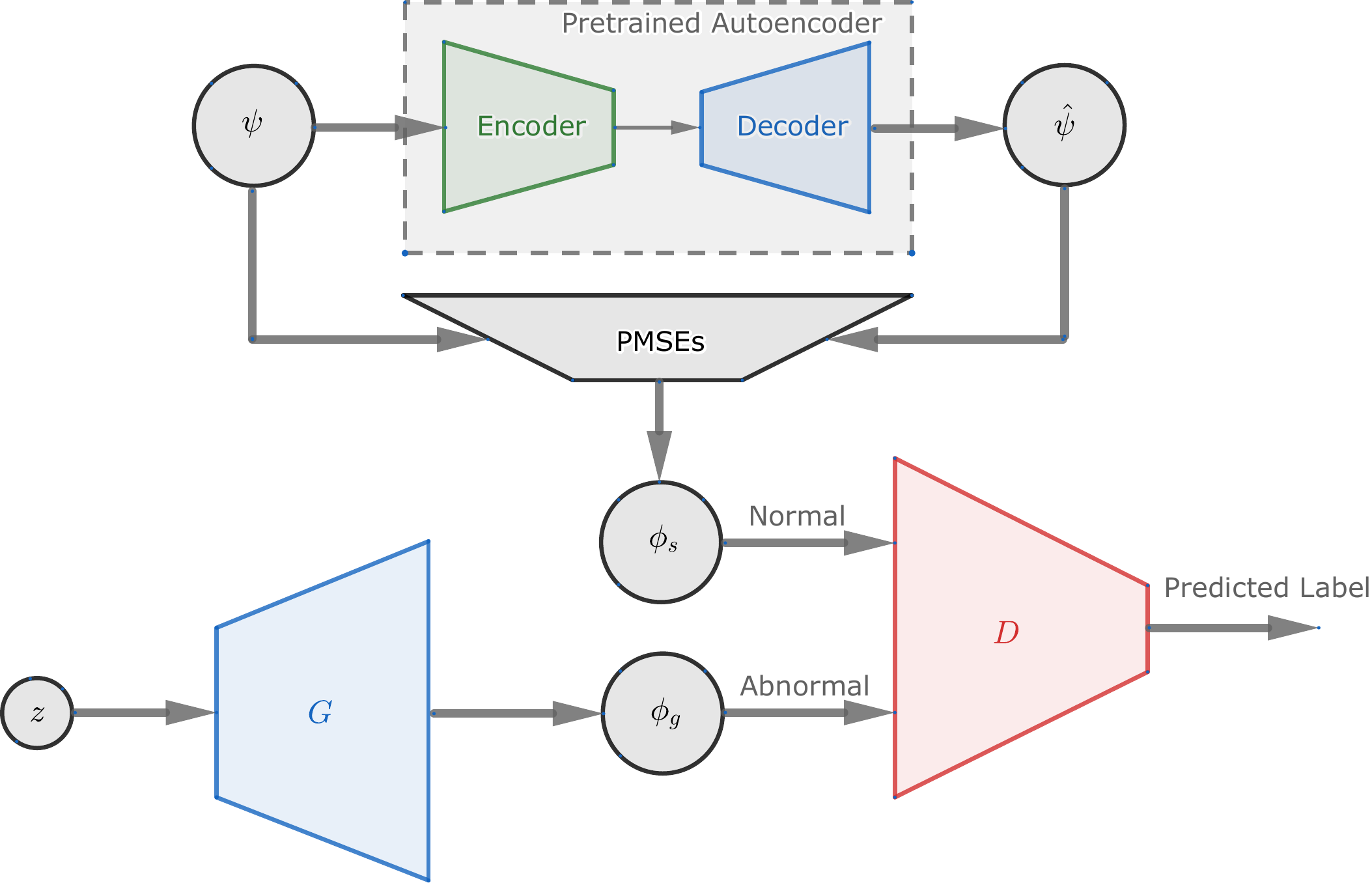}
    \caption{Architecture of our GAN-based approach for abnormal trajectory detection using a pretrained deep fully-connected AE.}
    \label{fig:ours_diag}
\end{figure}

Despite the recent success in using the GAN framework, it is very unstable for trajectory data structure when the generator $G$ is trained to generate realistic trajectories that are ordered sequences of points that must respect spatial and temporal constraints in order to fool the discriminator $D$. To tackle this problem, we propose to train $G$ to generate instead some trajectory reconstruction mean squared errors (MSEs) that reassemble the ones obtained by using a trained DAE with normal trajectory data. The MSE is obtained with normal data by calculating the error between the input and the output of the DAE. The model $D$ of our framework learns to distinguish real reconstruction MSEs from fake ones. Also, an adversarial criteria-based learning algorithm, which controls the learning speed of $D$ and stops the training as soon as a particular criterion is achieved, is followed during the training process of our network in order to let $D$ learn the most realistic abnormal MSEs as fake, while ensuring to classify the normal MSEs as real. When the training is completed, $D$ is used for predicting if the observed trajectory data is normal (real) or not (fake). Figure \ref{fig:ours_diag} illustrates the general architecture of our adversarial approach for detecting trajectory-based abnormalities.

\begin{figure}[t]
    \centering
    \includegraphics[width=0.74\linewidth]{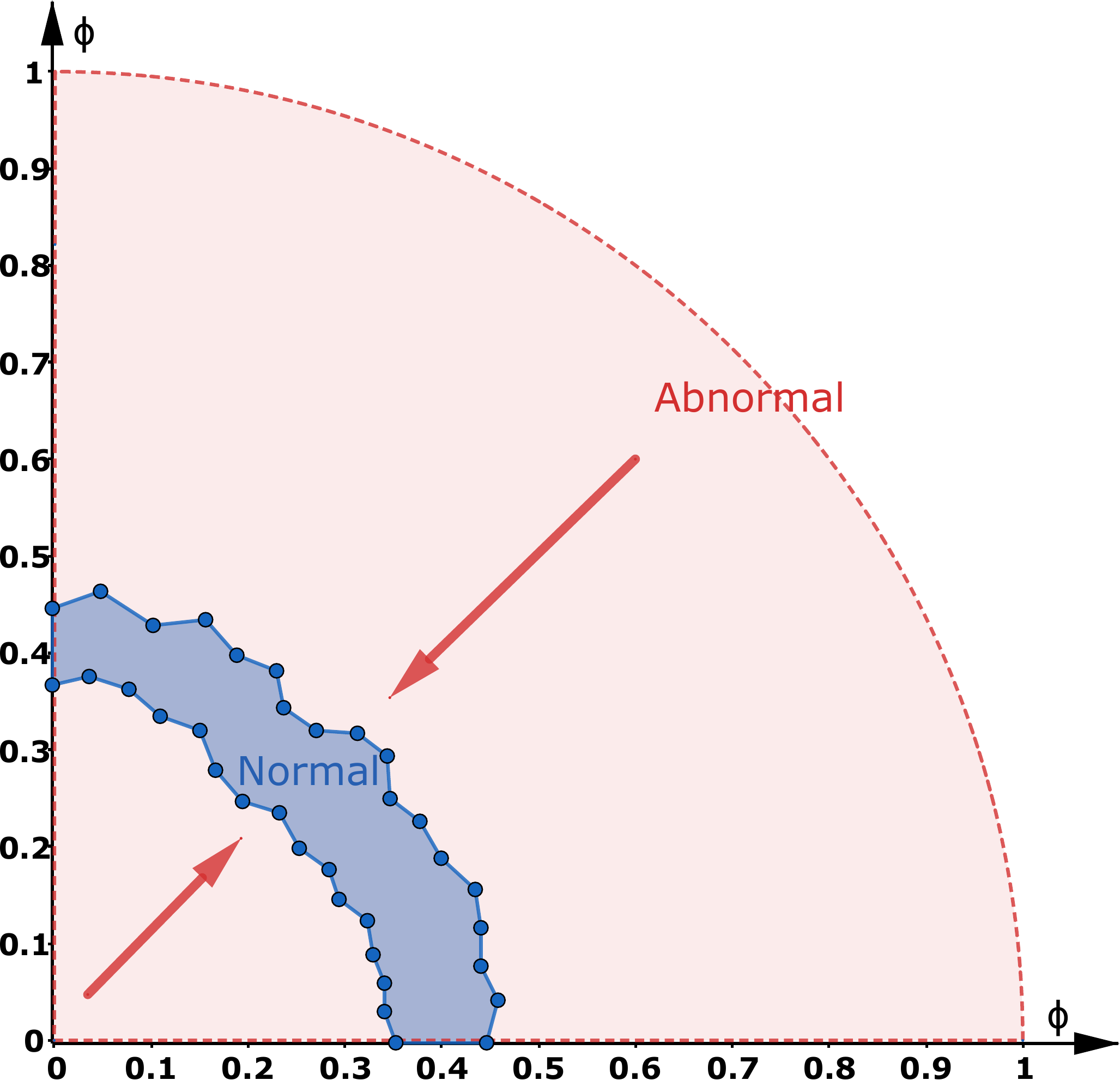}
    \caption{Example of the abnormal reconstruction MSE spectrum $\phi$ that the model $G$ roughly generates during the training. Normal MSEs are extracted by using the pretrained AE model. The axes represent the MSE values in which all the input data are normalized between 0 and 1. The model $D$ learns to distinguish the normal $\phi$ (illustrated by the area in between blue dots) from the abnormal one (in red). Red arrows illustrate that generated MSEs are getting closer to normal ones as the training of $G$ progresses.}
    \label{fig:normal_abnormal_border}
\end{figure}

The goal of our $G$ during the training process is to start from generating random sets of reconstruction MSEs and to finish with the most realistic MSEs that might still be considered as abnormal. For example, a car that drives in the wrong direction will result in a very similar looking reconstruction MSEs when using the pretrained DAE, but the $D$ model should be able to identify this as abnormal/fake. Figure \ref{fig:normal_abnormal_border} shows the spectrum of abnormality that the generator $G$ is approximately generating during the training process.

Before training our adversarial network, a DAE must be trained using the extracted and augmented trajectory data considered as normal, as already summarized above. After that, the partial averages of the reconstruction error ($\phi_s$) are extracted in the following way for each trajectory sample $s$:

\begin{align}
    \begin{split}
        \phi_s &= \resizebox{.8\hsize}{!}{$\left[ \frac{\sum\limits_{i=0}^{L-1} \left( \psi(i) - \hat{\psi}(i) \right)^2}{L}, \frac{\sum\limits_{i=L}^{2L-1} \left( \psi(i) - \hat{\psi}(i) \right)^2}{L}, \dots, \frac{\sum\limits_{i=(M-1) L}^{M L-1} \left( \psi(i) - \hat{\psi}(i) \right)^2}{L} \right]$} \\
        &= \left[ \textnormal{PMSE}_0, \textnormal{PMSE}_1, \dots, \textnormal{PMSE}_{M-1} \right] \textnormal{,}
    \end{split}
\end{align}

where PMSE is the partial mean squared error with respect to the input trajectory sample size $N = M \times L$. By using the PMSEs instead of one MSE per sample, $D$ can also learn the "normal" level of variations among each PMSE in $\phi_s$. This enables $D$ to distinguish the most realistic looking trajectory. Once the corresponding AE is trained and the $\phi_s$ are extracted for every training trajectory, the adversarial model, constituting of two networks $G$ and $D$, is trained following the objective function:

\begin{gather}
    \min\limits_{G} \max\limits_{D} \left( \resizebox{.7\hsize}{!}{$\EX_{\phi \sim p_\phi} \left[ \log D(\phi) \right] + \EX_{z \sim p_z} \left[1 - \log D(G(z)) \right]$} \right) \label{eq:obj_func_ours}
\end{gather}

where $p_\phi$ represents the distribution of $\phi_s$ retrieved using the pretrained autoencoder with the normal (real) data and $D(\dots)$ is the classification decision of the discriminator $D$.

Algorithm \ref{algo:gan_training} presents the training process that stabilizes the learning of both $G$ and $D$ models of the GAN framework by basically controlling the learning rate of $G$ ($\iota_g$) relative to that of $D$ ($\iota_d$) based on the performance of $G$. The training stops as soon as $D$ detects almost all the generated samples by $G$ as real or, when it reaches the maximum number of epochs $E_{max}$. In the learning algorithm, $\alpha_{D}(\phi_s)$ corresponds to the classification by $D$ of $\phi_s$ as real, $\alpha_{D}(\phi_g)$ is the detection of the generated $\phi_g$ as fake and $\alpha_{G}(\phi_g)$ represents the classification by $D$ of $\phi_g$ as real. Therefore, $\alpha_{D}(\phi_g) = 1 - \alpha_{G}(\phi_g)$.

\begin{algorithm}[t] 
\caption{Training of our Adversarial Classifier Model.} 
\label{algo:gan_training} 
\begin{algorithmic}[1]
\Require normalized normal trajectory data $\psi$
\State Initialize $\iota_g$ to $\iota_d / 2$
\State Set the minimum learning rate of $G$ to $\iota_d / 4$
\For {$e=1$ to $E_{max}$} \Comment{epoch}
    \State Generate random noise $z$
    \State Extract a random batch of input data $\psi_b$
    \State Generate a batch of new data $\phi_g$ using $G$
    \begin{equation*}
        \phi_g = G(z)
    \end{equation*}
    \State Train $D$ with batch $\psi_b$ labelled as valid $ = 1$
    \begin{algsubstates}
        \State Reconstruct $\hat{\psi}_b$ using the pretrained DAE
        \State Compute the squared errors $( \psi_b - \hat{\psi}_b )^2$
        \State Compute $\phi_s$
        \State Fit $D$ using $\phi_s$ and get the accuracy $\alpha_{D}(\phi_s)$
    \end{algsubstates}
    \If {$\alpha_{D}(\phi_s) \geq 0.99 $}
    \State \multiline{Train $D$ using $\phi_g$ with label '0' and get the accuracy value $\alpha_{D}(\phi_g)$}
    \State \multiline{Train $G$ by fitting the combined model ($G + D$) using $z$ with label '1' while keeping the weights of $D$ constant and store the accuracy $\alpha_{G}(\phi_g)$}
    \State Generate new $\phi_g$ using $G$
    \State Adjust $\iota_g$ depending on $\alpha_{G}(\phi_g)$
    \begin{algsubstates}
        \State $\iota_g$ varies from $\iota_d / 2$ to $\iota_d / 4$
        \State When $\alpha_{G}(\phi_g)$ gets higher, $\iota_g$ becomes smaller
    \end{algsubstates}
    \If {$K$ consecutive values of $ \alpha_{G}(\phi_g) > 0.95 $}
    \State Stop the training.
    \EndIf
    \EndIf
    \State \multiline{Test $D$ using the new $\phi_g$ and get the global accuracy $\alpha_D = (\alpha_{D}(\phi_s) + \alpha_{D}(\phi_g))/2 $}
\EndFor
\end{algorithmic}
\end{algorithm}

As previously mentioned, only the discriminator $D$ network is used for the testing phase of our GAN framework. Also, the proposed adversarial approach can be seen as a way of transforming an unsupervised network, a DAE in our case, into a supervised binary classifier without the need of using abnormal data.

\section{Experiments}
\begin{figure*}[t]
    \centering
	\subfloat[Sherbrooke: normal\label{fig:traj_dataset_sb_n}]{%
        \includegraphics[height = \trajfigheight\linewidth,width = \trajfigwidth\textwidth]{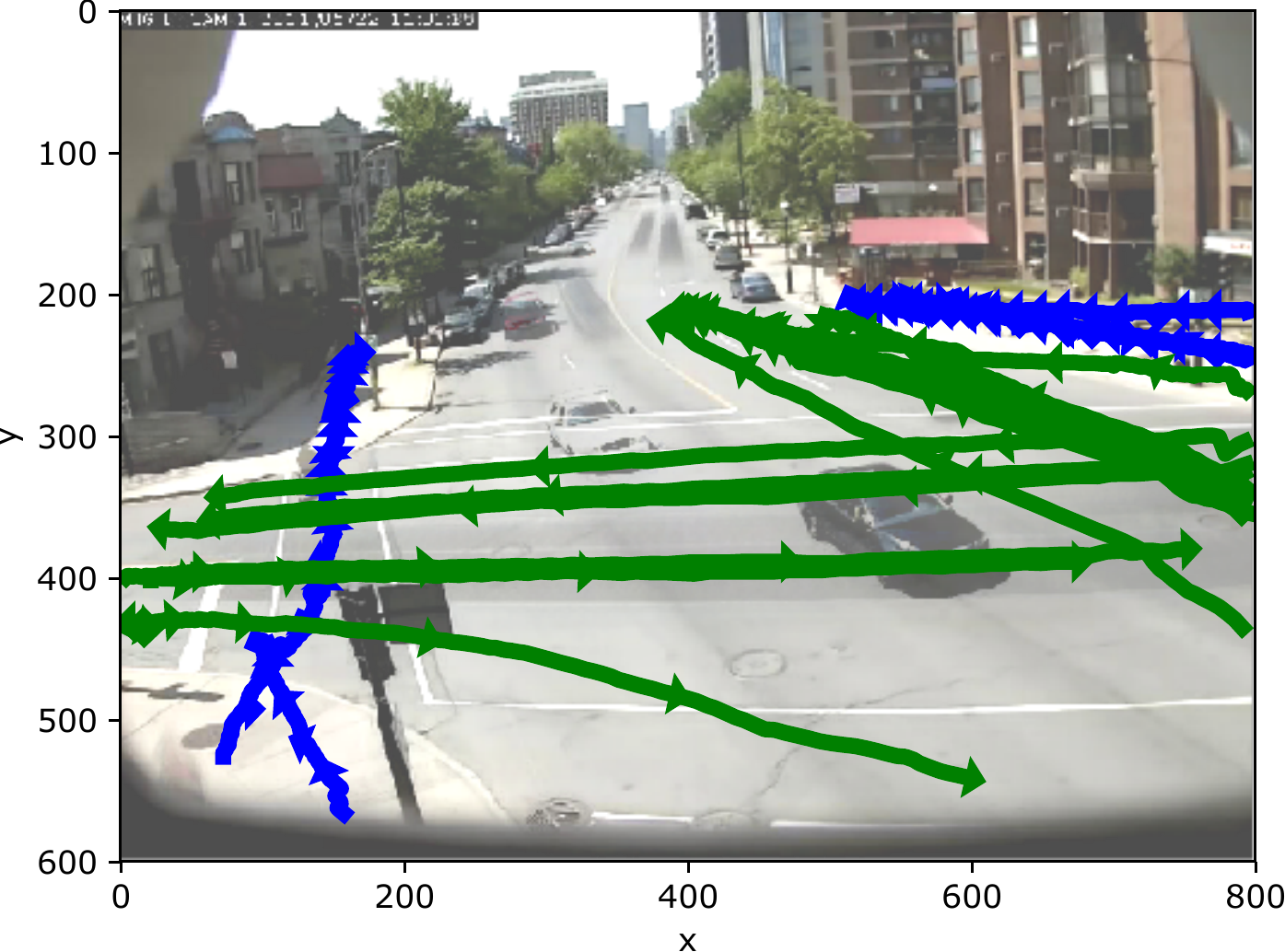}
    }
    \subfloat[Rouen: normal\label{fig:traj_dataset_ru_n}]{%
        \includegraphics[height = \trajfigheight\textwidth,width = \trajfigwidth\textwidth]{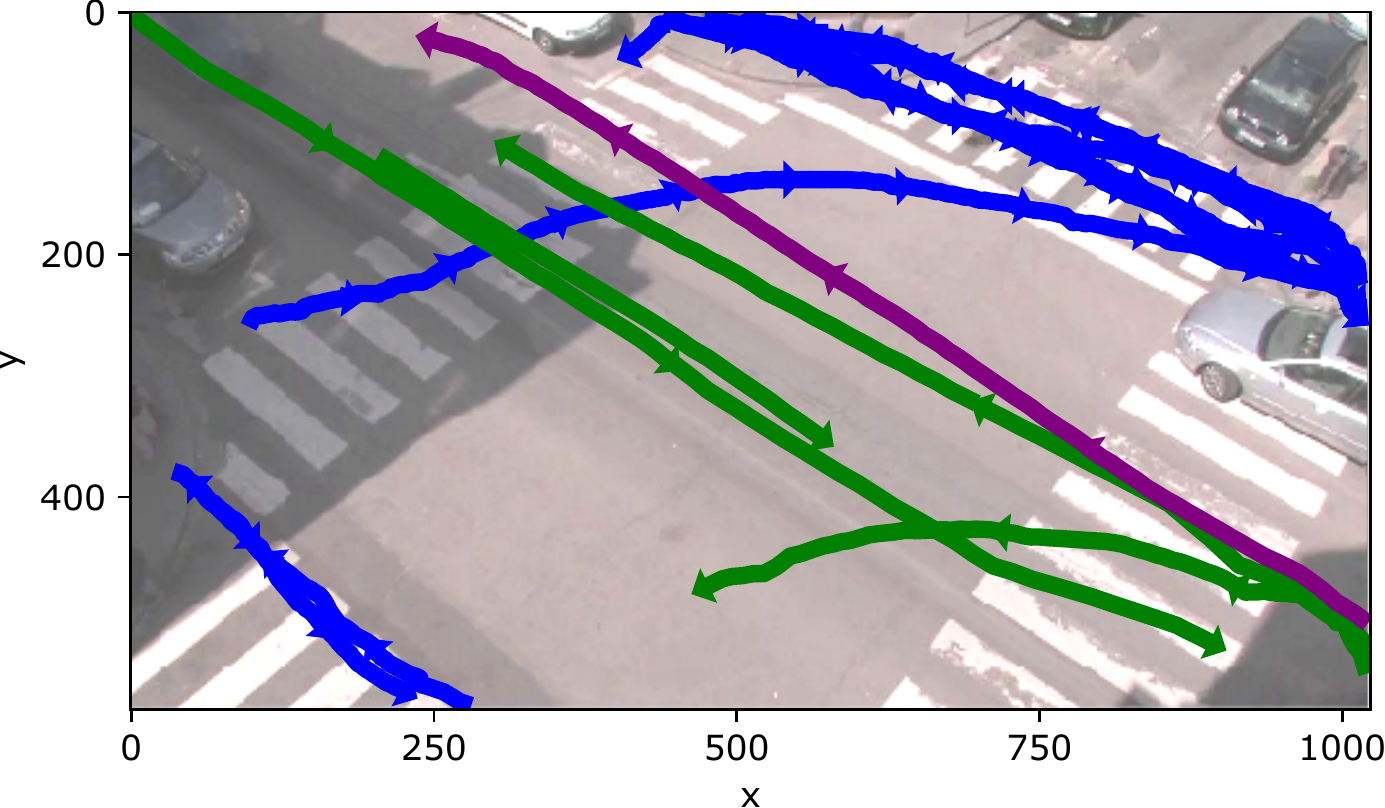}
    }
    \subfloat[St-Marc: normal\label{fig:traj_dataset_sm_n}]{%
        \includegraphics[height = \trajfigheight\textwidth,width = \trajfigwidth\textwidth]{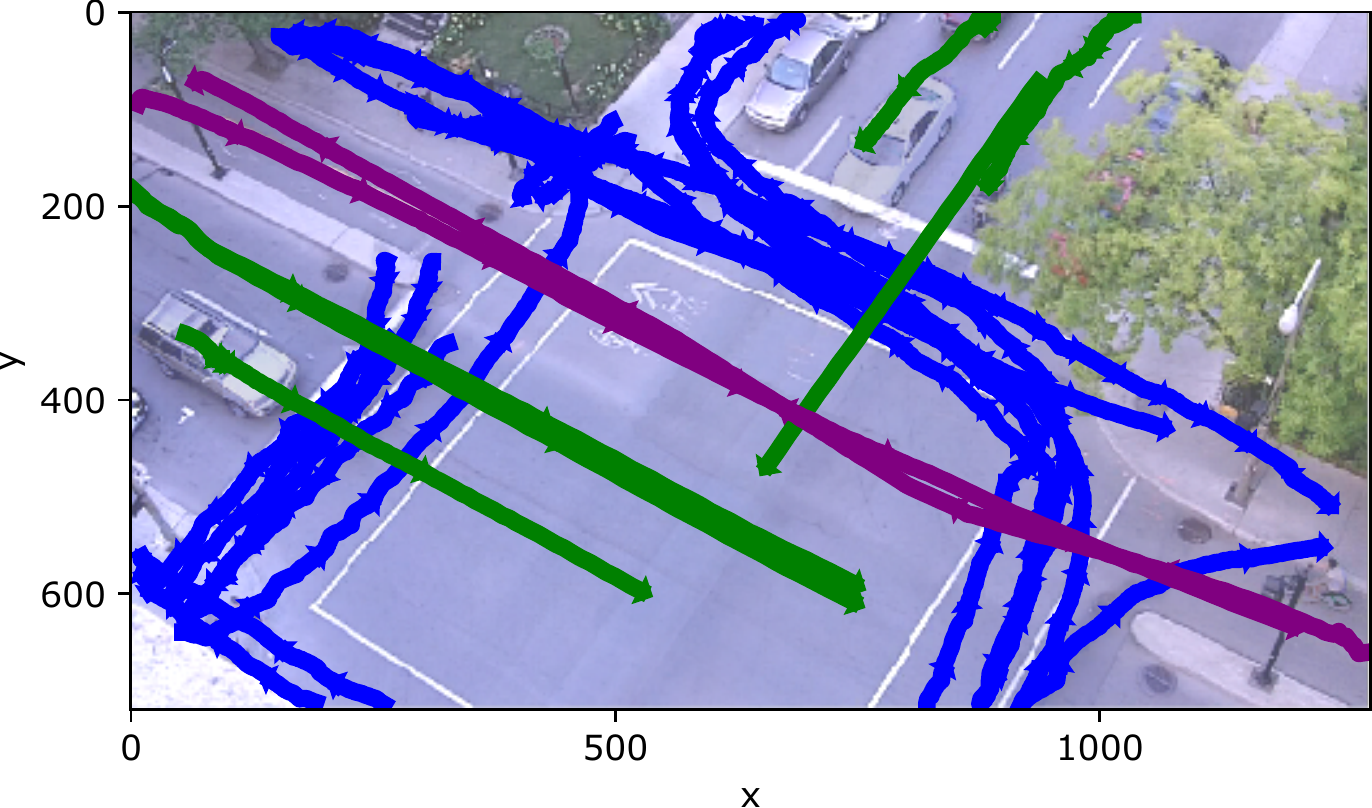}
    }
    \subfloat[René-Lévesque: normal\label{fig:traj_dataset_rl_n}]{%
        \includegraphics[height = \trajfigheight\textwidth,width = \trajfigwidth\textwidth]{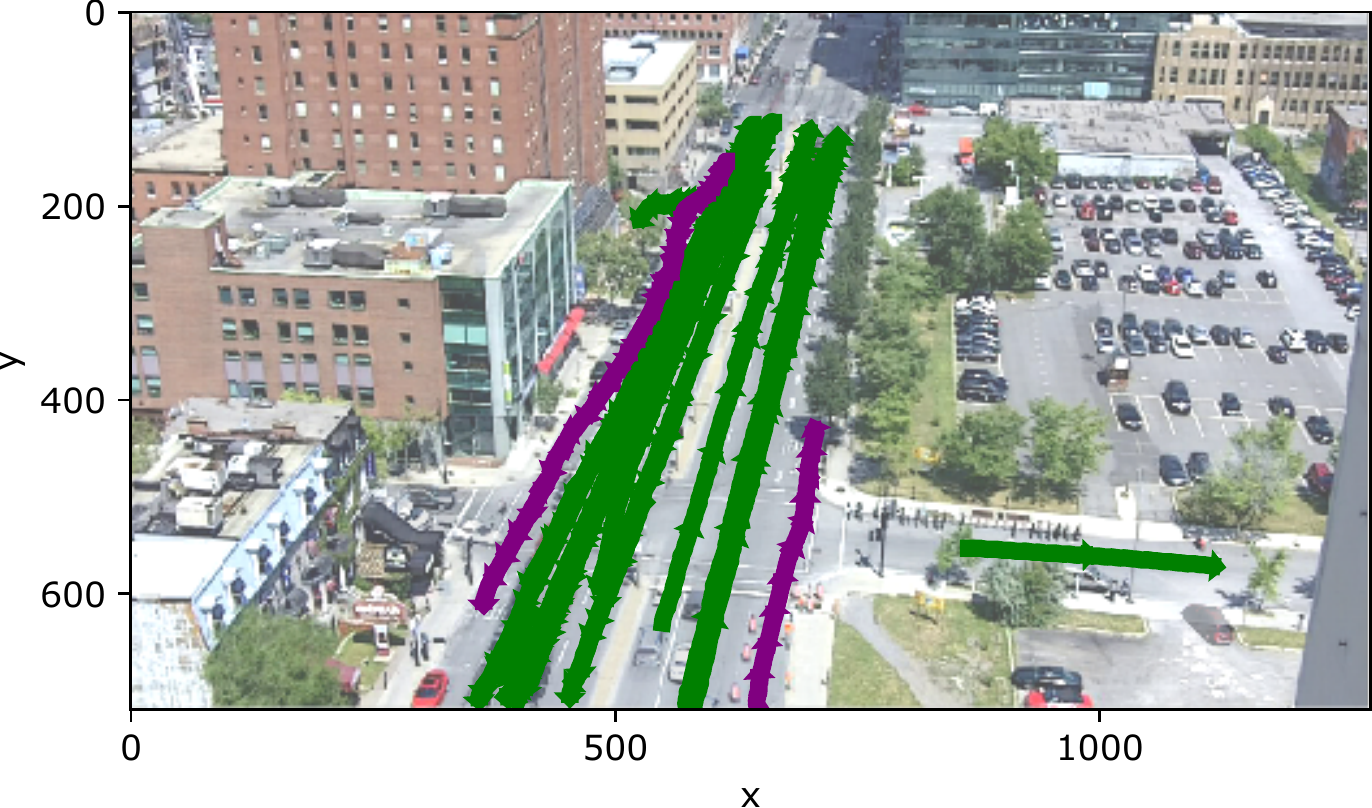}
    }
    \\ \vspace{-0.8\baselineskip}
    \subfloat[Sherbrooke: abnormal\label{fig:traj_dataset_sb_an}]{%
        \includegraphics[height = \trajfigheight\linewidth,width = \trajfigwidth\textwidth]{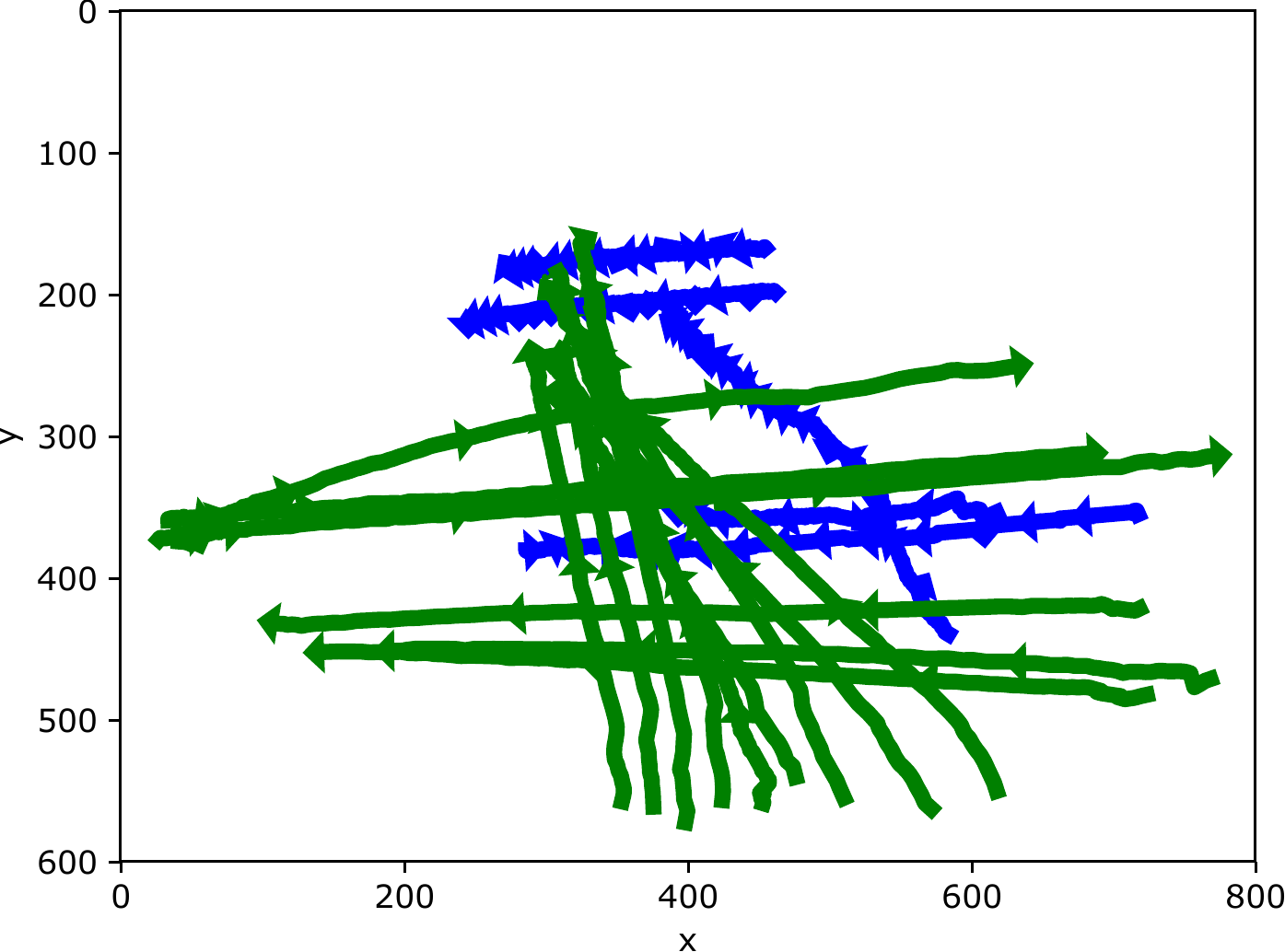}
    }
    \subfloat[Rouen: abnormal\label{fig:traj_dataset_ru_an}]{%
        \includegraphics[height = \trajfigheight\textwidth,width = \trajfigwidth\textwidth]{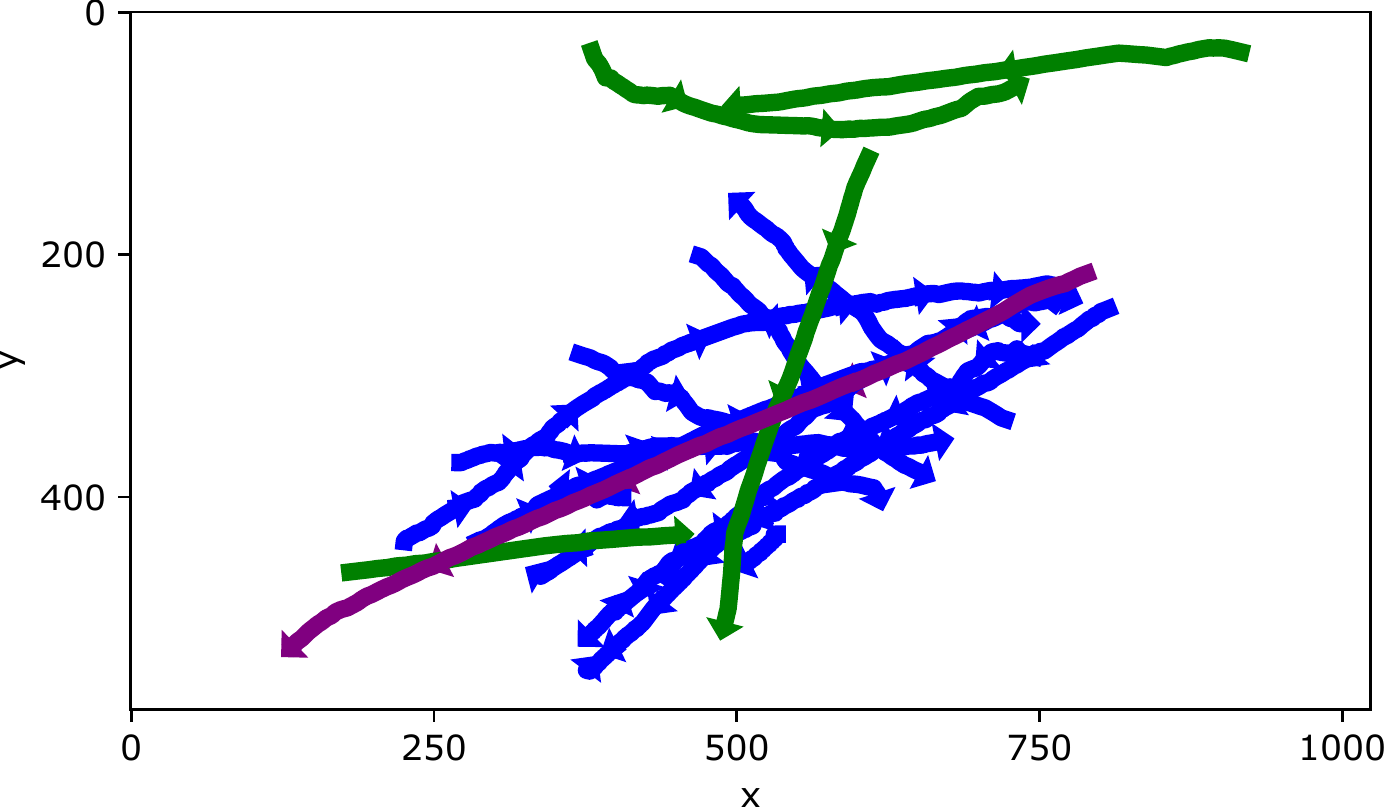}
    }
    \subfloat[St-Marc: abnormal\label{fig:traj_dataset_sm_an}]{%
        \includegraphics[height = \trajfigheight\textwidth,width = \trajfigwidth\textwidth]{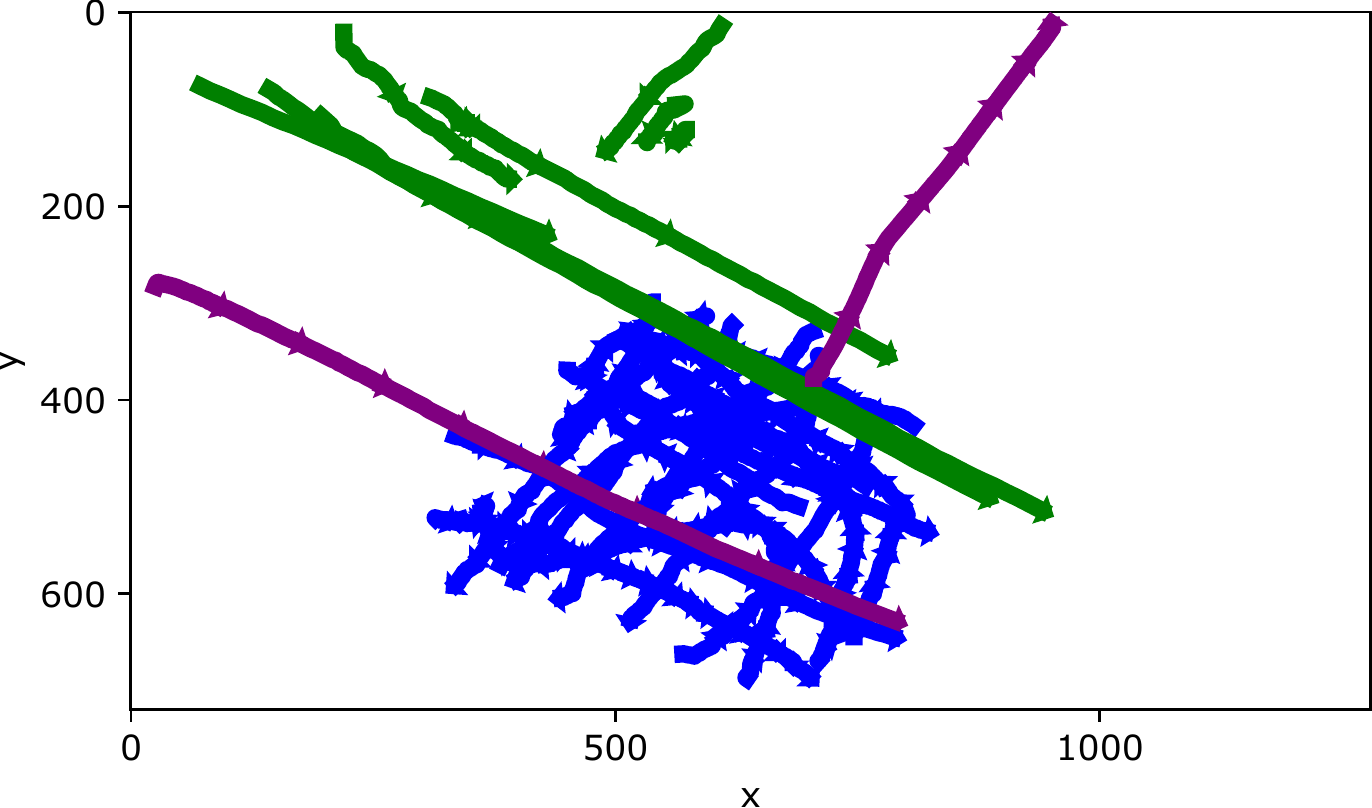}
    }
    \subfloat[René-Lévesque: abnormal\label{fig:traj_dataset_rl_an}]{%
        \includegraphics[height = \trajfigheight\textwidth,width = \trajfigwidth\textwidth]{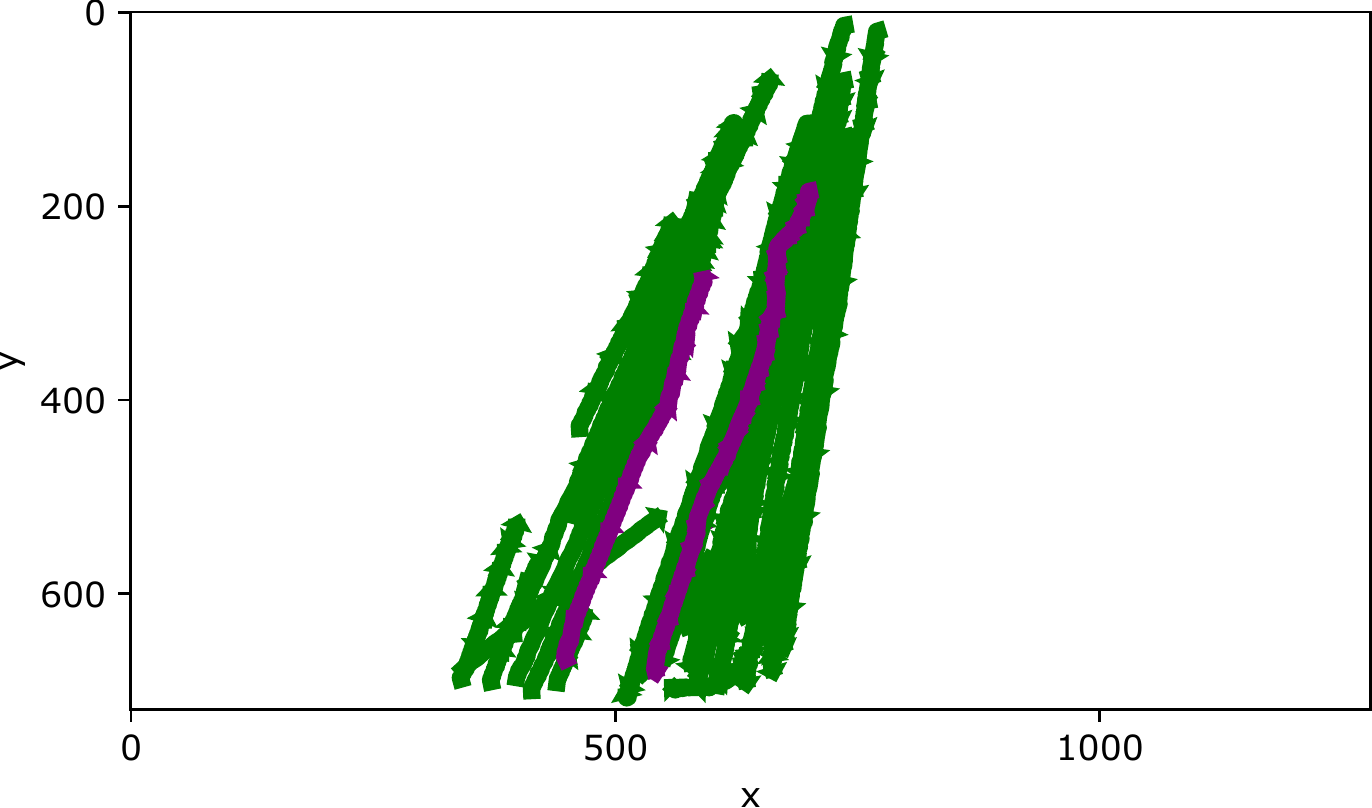}
    }
    \caption{Observed (normal) and created (abnormal) trajectories of the Urban Tracker Dataset. The coloured arrows illustrate trajectory direction where the road users are represented by colours: blue: pedestrians, green: cars, and purple: bikes.}
    \label{fig:traj_dataset}
\end{figure*}

Our proposed adversarial method for classifying abnormal trajectories (ALREC) is tested on four different outdoor urban traffic videos obtained from the Urban Tracker dataset \cite{urban_tracker}, in which 1) the Sherbrooke video includes 15 cars and 5 pedestrians, 2) the Rouen video includes 4 vehicles, 1 bike and 11 pedestrians, 3) the St-Marc video includes 7 vehicles, 2 bicycles and 19 pedestrians and 4) the René-Lévesque video contains 29 vehicles and 2 bikes. As in \cite{article1}, all the observed trajectories are considered as being normal. To test our model, abnormal trajectories were generated. They correspond to 1) pedestrians walking in the middle of the road, 2) vehicles driving on the opposite/wrong directions, or in the bike paths or on the sidewalks, and 3) bikes going outside the bike paths when there is one. Figure \ref{fig:traj_dataset} illustrates normal and abnormal trajectories.

Also, in order to demonstrate the genericity of our proposed method, we conducted ablation studies in which ALREC is applied to six scenes from the INRIA dataset of the CAVIAR project \cite{ref_caviar}. As in \cite{ref_article_traditional_trajectory_method}, we applied our method on 40 scenarios where 21 are labelled as normal (person walking, browsing, resting, people meeting) and 19 are labelled as abnormal (person slumping or fainting, leaving bags behind and people fighting).

\subsection{Experimental Setup}
For comparing ALREC to the method of \cite{article1}, the same hyperparameters are used during the training of the DAE. To summarize it briefly, the input size of a trajectory sample is set to $(1 + 4 \times m) = 125$, with $m = 31$ referring to the section \ref{sec: 3.1}. Then, we train the DAE, containing the following number of neurons in the hidden layers from the first to the last CFV layer: 128, 64, 32, 16, 8, with a batch size of 128 for 100 epochs. The learning rate is fixed to $0.001$ using the RMSProp optimizer and the loss function is based on the MSE as presented in equation \ref{eq:obj_func_ae}. The trajectory samples are collected by sliding a window over complete trajectories, and each complete trajectory of an object is augmented by randomly generating positions and velocities near the original ones.

The additional parameters necessary for the training of ALREC are presented in table \ref{tab:gan_params}. We have noticed that the discriminator learns better to differentiate between normal and abnormal PMSEs when the vector size of $\phi_s$ and $\phi_g$ is small enough and therefore $M = 5$ seems convenient because of the fact that $N$ is divisible by $M$. For building the networks $G$ and $D$, each layer has a number of neurons that is a power of two except for the input and the output layers in which they must have the same sizes of the desired input and output. Note that the size of the input layer in the $G$ network is the same as the size of each random sample $z$. For the model $D$, which is the crucial part of our proposed method, we added an extra hidden layer in order to avoid over-fitting. We use the RMSprop optimizer during the training of our networks and the learning rate of $G$ varies following algorithm \ref{algo:gan_training}, where $K = 100$. Also, the ReLU and the Sigmoid activation are used in the hidden and the output layers of all our networks respectively. Note that all the implementations\footnote{Codes in \url{https://github.com/proy3/Abnormal_Trajectory_Classifier}.} are done in a Python environment using the Keras package \cite{ref_chollet2015keras}.

\begin{table}[t]
\centering
\caption{Parameters and hyperparameters used for the training of ALREC.}\label{tab:gan_params}
\begin{tabular}{|c|c|l|}
\hline
Label &  Value & Definition\\
\hline \hline
$N$ &  125 & Size of each input sample \\
$M$ &  5 & Size of $\phi_s$ and $\phi_g$ \\
$G_{in}$ &  8 & Size of input layer of $G$\\
$G_{h_1}$ &  128 & Size of hidden layer 1 of $G$\\
$G_{h_2}$ &  64 & Size of hidden layer 2 of $G$\\
$G_{h_3}$ &  32 & Size of hidden layer 3 of $G$\\
$G_{h_4}$ &  16 & Size of hidden layer 4 of $G$\\
$G_{out}$ &  5 & Size of output layer of $G$\\
$D_{in}$ &  5 & Size of input layer of $D$\\
$D_{h_1}$ &  128 & Size of hidden layer 1 of $D$\\
$D_{h_2}$ &  64 & Size of hidden layer 2 of $D$\\
$D_{h_3}$ &  32 & Size of hidden layer 3 of $D$\\
$D_{h_4}$ &  16 & Size of hidden layer 4 of $D$\\
$D_{h_5}$ &  8 & Size of hidden layer 5 of $D$\\
$D_{out}$ &  1 & Size of output layer of $D$\\
$S_{bat}$ &  128 & Batch size for $G$ and $D$\\
$E_{max}$ &  5e+6 & Maximum number of epochs\\
$\iota_d$ & 5e-6 & Learning rate of $D$ \\
$\Theta_d$ & $\RMSprop\left( \iota_d \right)$ & Optimizer RMSprop for $D$ \\
$\Theta_g$ & $\RMSprop\left( \iota_g \right)$ & Optimizer RMSprop for $G$ \\
\hline
\end{tabular}
\end{table}

\subsection{Evaluation Procedure} \label{sec:eval_proc}
We follow the same evaluation procedure as in \cite{article1}. Once ALREC is fully trained using 80\% of the normal trajectory dataset, it is tested with the other 20\% of normal data and the abnormal data. The evaluation of the accuracy is based on Normal Detection Accuracy (NDA) and Abnormal Detection Accuracy (ADA). Given the number of normal sample $Size_N$ and the number of abnormal sample $Size_A$, NDA and ADA are given by $\textnormal{NDA} = \frac{Size_N}{Size_N+Size_A}$ and $\textnormal{ADA} = \frac{Size_A}{Size_N+Size_A}$.

\subsection{Results}

\subsubsection{Abnormal Trajectory Detection}
For demonstrating that our proposed adversarial classifier (ALREC) performs better than other well-known outliers detection methods, like One-Class Support Vector Machine (OC-SVM) \cite{ref_Scholkopf} and Isolation Forest (IF) \cite{ref_isolationforest}, and two other based on a single layer AE and an DAE \cite{article1}, we trained the previous methods (using the code provided by the authors of \cite{article1}) with the same normal and abnormal trajectory data. Table \ref{tab:results} shows the results for the classification of each trajectory sample. Note that, although we tried many adversarial approaches proposed in \cite{ref_Adversarial_Autoencoders,ref_Adversarially_Classifier_Sabokrou,ref_BiGAN,ref_social_GAN}, our ALREC is the first adversarial model that is compatible with the structure of our trajectory sample, where the training of the generator $G$ and the discriminator $D$ are stable.

\begin{table*}[t]
\centering
\caption{Abnormal and normal road user trajectory detection results. Boldface values imply the best precision. Data: name of video, Type: road user type, OC-SVM: One-Class SVM, IF: Isolation Forest, AE: simple AE, DAE: Deep AE \cite{article1}, ALREC: Our method.}\label{tab:results}
\begin{tabular}{|c|c|c|c||c|c|c|c|c|c|c|c|c|c|}
\multicolumn{4}{c}{} & \multicolumn{10}{c}{Model} \\ \cline{5-14}
\multicolumn{4}{c}{} & \multicolumn{2}{|c|}{OC-SVM} & \multicolumn{2}{c|}{IF} & \multicolumn{2}{c|}{AE} & \multicolumn{2}{c|}{DAE \cite{article1}} & \multicolumn{2}{c|}{ALREC (ours)} \\ \hline
Data & Type & Size$_N$ & Size$_A$ & \textnormal{ADA} & \textnormal{NDA} & \textnormal{ADA} & \textnormal{NDA} & \textnormal{ADA} & \textnormal{NDA} & \textnormal{ADA} & \textnormal{NDA} & \textnormal{ADA} & \textnormal{NDA} \\ \hline \hline
\multicolumn{1}{|c|}{\multirow{3}{*}{Sherb.}} & Cars & 8824 & 174 & \textnormal{0.56} & \textnormal{0.88} & 0.63 & \textnormal{0.80} & \textnormal{0.26} & \textbf{0.99} & \textnormal{0.83} & \textbf{0.99} & \textbf{0.95} & \textbf{0.99} \\
\multicolumn{1}{|c|}{} & Peds & 11782 & 232 & \textnormal{0.23} & \textnormal{0.91} & 0.00 & \textnormal{0.99} & \textnormal{0.06} & \textbf{1.00} & \textnormal{0.82} & \textbf{1.00} & \textbf{0.96} & \textnormal{0.99} \\
\multicolumn{1}{|c|}{} & All & 20606 & 406 & \textnormal{0.37} & \textnormal{0.90} & \textnormal{0.28} & 0.89 & \textnormal{0.13} & 0.99 & \textnormal{0.82} & \textbf{1.00} & \textbf{0.96} & \textnormal{0.99} \\
\hline
\multicolumn{1}{|c|}{\multirow{4}{*}{Rouen}} & Cars & 1429 & 29 & \textnormal{0.48} & \textnormal{0.54} & \textnormal{0.55} & 0.26 & \textnormal{0.31} & \textbf{1.00} & \textnormal{0.58} & \textbf{1.00} & \textbf{0.90} & \textbf{1.00} \\
\multicolumn{1}{|c|}{} & Peds & 9843 & 193 & \textnormal{0.00} & \textnormal{0.95} & \textnormal{0.00} & \textbf{1.00} & \textnormal{0.00} & \textbf{1.00} & \textnormal{0.98} & \textbf{1.00} & \textbf{0.99} & \textbf{1.00} \\
\multicolumn{1}{|c|}{} & Bike & 612 & 12 & \textnormal{0.00} & \textnormal{0.86} & \textnormal{0.33} & 0.81 & \textnormal{0.00} & \textbf{1.00} & \textbf{1.00} & \textbf{1.00} & \textbf{1.00} & \textbf{1.00} \\
\multicolumn{1}{|c|}{} & All & 11884 & 234 & \textnormal{0.05} & \textnormal{0.90} & \textnormal{0.07} & 0.89 & \textnormal{0.03} & \textbf{1.00} & \textnormal{0.93} & \textbf{1.00} & \textbf{0.96} & \textbf{1.00} \\
\hline
\multicolumn{1}{|c|}{\multirow{4}{*}{St-Marc}} & Cars & 4488 & 88 & \textnormal{0.07} & \textnormal{0.68} & 0.31 & \textnormal{0.62} & \textnormal{0.03} & \textbf{0.96} & \textnormal{0.52} & \textbf{0.96} & \textbf{0.67} & \textbf{0.96} \\
\multicolumn{1}{|c|}{} & Peds & 33407 & 657 & \textnormal{0.00} & \textnormal{0.94} & \textnormal{0.00} & \textnormal{0.98} & \textnormal{0.00} & \textbf{1.00} & \textnormal{0.76} & \textbf{1.00} & \textbf{0.94} & \textbf{1.00} \\
\multicolumn{1}{|c|}{} & Bike & 2244 & 44 & \textnormal{0.11} & \textnormal{0.72} & 0.61 & \textnormal{0.41} & \textnormal{0.02} & \textbf{1.00} & \textnormal{0.72} & \textbf{1.00} & \textbf{0.93} & \textbf{1.00} \\
\multicolumn{1}{|c|}{} & All & 40139 & 789 & \textnormal{0.01} & \textnormal{0.90} & \textnormal{0.07} & \textnormal{0.90} & \textnormal{0.00} & \textbf{0.99} & \textnormal{0.67} & \textbf{0.99} & \textbf{0.92} & \textbf{0.99} \\
\hline
\multicolumn{1}{|c|}{\multirow{3}{*}{Rene-L.}} & Cars & 38762 & 762 & \textnormal{0.16} & \textnormal{0.91} & \textnormal{0.29} & \textnormal{0.90} & \textnormal{0.00} & \textbf{0.99} & \textnormal{0.30} & \textbf{0.99} & \textbf{0.61} & \textbf{0.99} \\
\multicolumn{1}{|c|}{} & Bike & 6579 & 129 & \textnormal{0.06} & \textnormal{0.84} & \textnormal{0.08} & 0.89 & \textnormal{0.00} & \textbf{1.00} & \textnormal{0.98} & \textbf{1.00} & \textbf{1.00} & \textbf{1.00} \\
\multicolumn{1}{|c|}{} & All & 45341 & 891 & 0.14 & \textnormal{0.90} & \textnormal{0.25} & 0.89 & \textnormal{0.00} & \textbf{0.99} & \textnormal{0.35} & \textbf{0.99} & \textbf{0.62} & \textbf{0.99} \\
\hline
\end{tabular}
\end{table*}

With our proposed ALREC, table \ref{tab:results} shows that the ADA is significantly better than all the other tested models including OC-SVM, IF, AE and DAE \cite{article1}, where the latter is used as the pretrained model in our ALREC. Therefore, we show that by adversarially learning to classify the reconstruction error extracted through the pretrained DAE, the abnormal trajectories are better detected, even with the most difficult ones of Rene-L. video. We notice that, generally, the ADA are noticeably better by around 20-30\% for St-Marc and Rene-L. videos in which the abnormal samples are very close to the normal ones.

However, for the Sherb. video, the NDA results with ALREC are 1\% lower than DAE \cite{article1}. The reason behind it is that our proposed ALREC model strictly follows the training algorithm \ref{algo:gan_training} in which the performance of NDA is limited by the condition that ensures to train $D$ with the generated samples only if 99\% of the normal reconstruction errors were detected as normal. Therefore, the "normal" generated samples by $G$, near the end of the training of ALREC, have a chance of lowering NDA by around 1\%. In our case, this minimal degradation in NDA is not an issue, mainly because of the fact that we assumed all the observed videos as "normal" even though there might be some that are not. In that logic, a model that is trained to focus slightly more on the detection of abnormal trajectories than the normal ones is more preferable.

\subsubsection{Abnormal Events Detection}
In this section, we demonstrate that, by using the INRIA lab entrance dataset of the CAVIAR project \cite{ref_caviar}, our proposed method can be generalized for other abnormal event detection and can give significantly better results than the state-of-the-art method of \cite{ref_article_traditional_trajectory_method}. Each trajectory sample, in this case, follows a slightly different structure than the one defined in equation \ref{eq:traj_struct}, in which the height $h$, the width $w$ and the orientation $o$ of each bounding box, which is already given in the ground truth, are taken into account.

We also added one additional hidden layer in the DAE and the GAN models, and adjusted the parameters of the table \ref{tab:gan_params} with $N = 232$, $M = 8$ and an additional hidden layer with $256$ units, which is added in between the input layer and the first hidden layer of the generator and the discriminator of the GAN framework, and also in the encoder and decoder of the DAE network.

Table \ref{tab:results_2} reports the results of ALREC compared to sparse reconstruction analysis (SRA) from \cite{ref_article_traditional_trajectory_method} by using Detection Accuracy (DACC: the number of normal and abnormal test samples detected as normal and abnormal respectively) and Correct Classification Rate (CCR: the number of correctly detected normal test samples as normal) metrics introduced in \cite{ref_article_traditional_trajectory_method}. Note that, in the case of ALREC, we consider that the complete trajectory of a user is abnormal if at least 5\% of the corresponding trajectory samples are abnormal. Also, we used the best results that were presented in \cite{ref_article_traditional_trajectory_method} with the same evaluation metrics in order to compare with their best results.

\begin{table}[t]
\centering
\caption{Behavioural event detection results with the INRIA dataset \cite{ref_caviar}. SRA: Method of   \cite{ref_article_traditional_trajectory_method}, ALREC: Our method, DACC: Detection ACCuracy \cite{ref_article_traditional_trajectory_method}, CCR: Correct Classification Rate \cite{ref_article_traditional_trajectory_method}.}\label{tab:results_2}
\begin{tabular}{c|c|c|}
 & SRA \cite{ref_article_traditional_trajectory_method} & ALREC (ours) \\ \hline
DACC \% & 90.42 & \textbf{100.00} \\ \hline
CCR \% & 70.09 & \textbf{100.00} \\ \hline
\end{tabular}
\end{table}

Results show that ALREC is able to detect all abnormal events as abnormal while detecting the normal ones as normal. Also, the improvement compared to SRA \cite{ref_article_traditional_trajectory_method} is quite significant with around 10\% in DACC and 30\% in CCR. This illustrates that our approach is not only limited to abnormal road user trajectories but can also be generalized for broader applications like detecting abnormal events/activities including people fighting, running or leaving bags behind.

\subsubsection{Ablation Study}
\begin{figure}[t]
    \centering
	\subfloat[ADA curves\label{fig:ablation_study_ADA}]{%
        \includegraphics[width=1\linewidth]{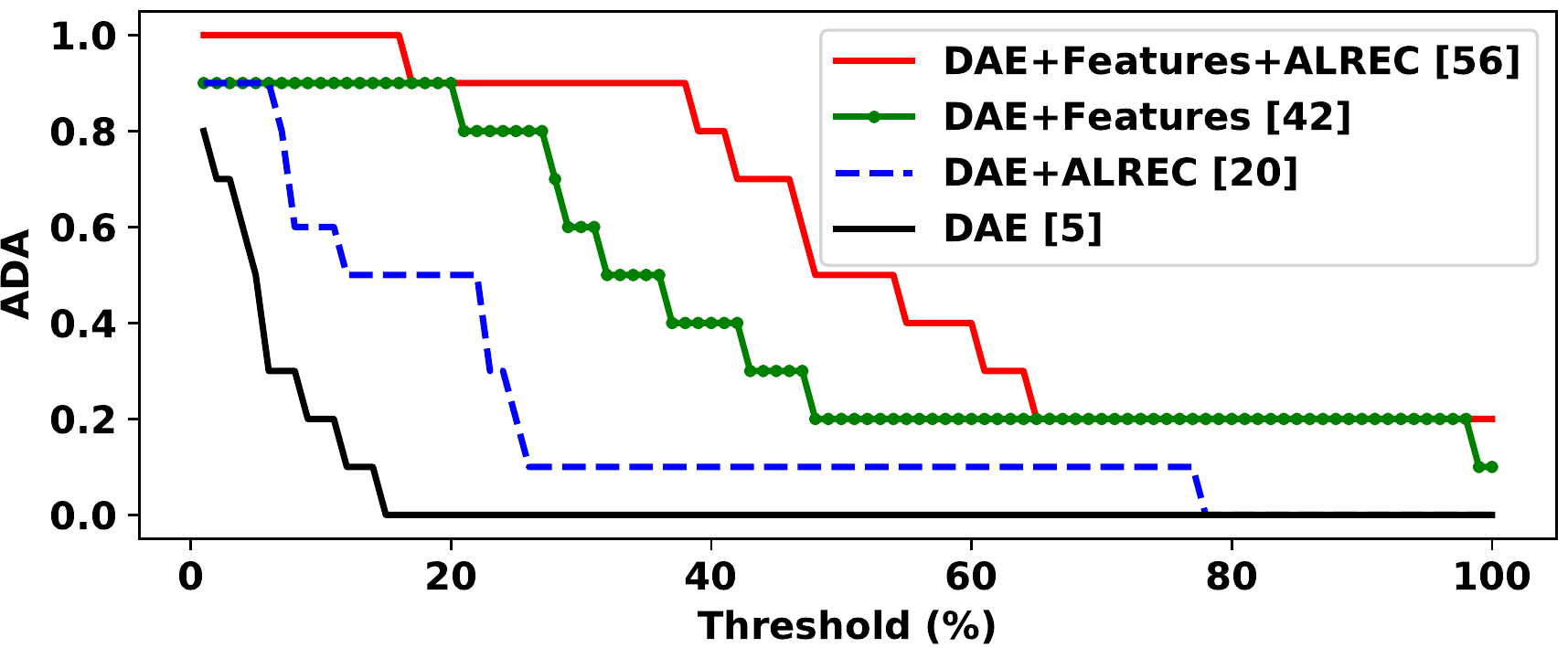}
    }
    \\ \vspace{-0.8\baselineskip}
    \subfloat[NDA curves\label{fig:ablation_study_NDA}]{%
        \includegraphics[width=1\linewidth]{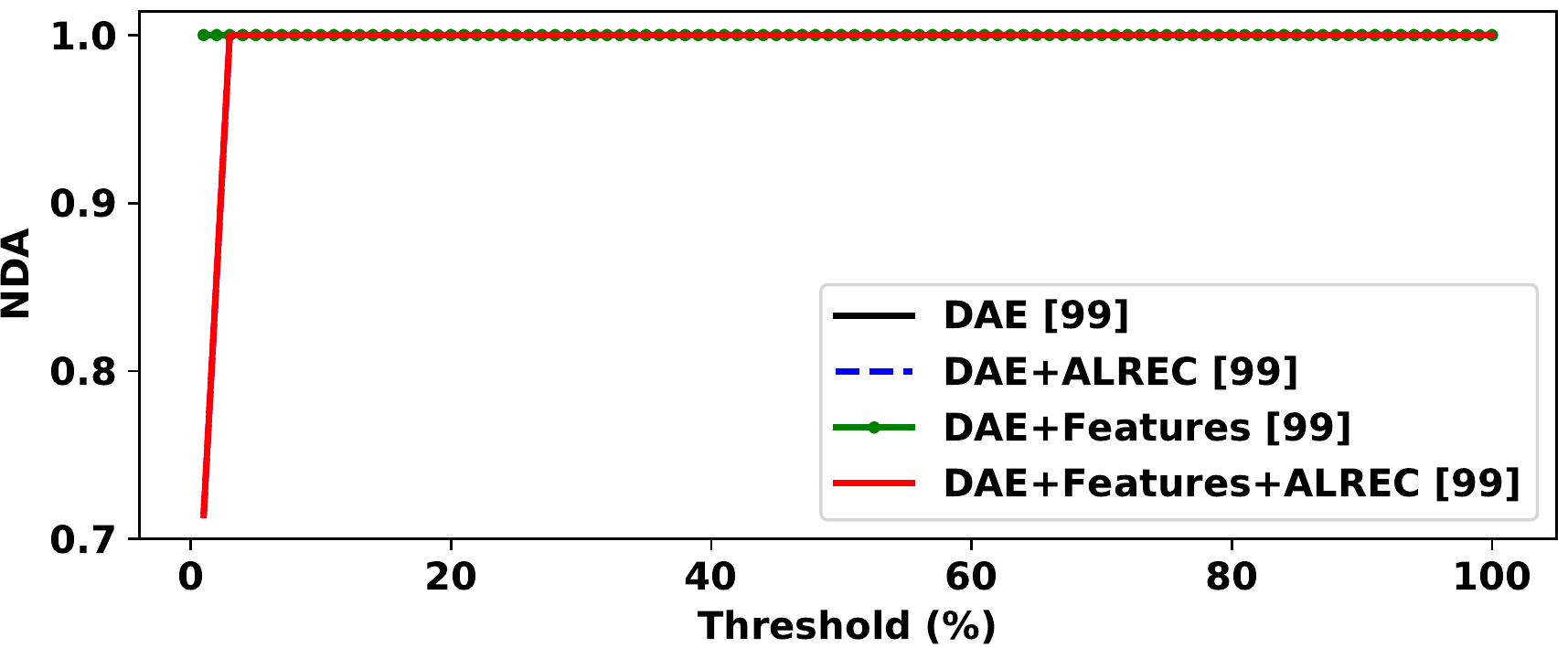}
    }
    \caption{Ablation study results based on ADA and NDA of complete trajectories with the INRIA dataset \cite{ref_caviar}. Features: height, width and orientation of the bounding box as additional features, Threshold: minimal percentage of trajectory samples of a complete trajectory of an object that needs to be detected as abnormal in order for the complete trajectory to be abnormal, [AUC]: area under the curve.}
    \label{fig:ablation_study}
\end{figure}

In order to further demonstrate the effectiveness of ALREC, we conducted an ablation study for the behavioural detection of complete trajectories using the INRIA lab entrance dataset in which we tested the various possible combination of DAE with ALREC are shown in figure \ref{fig:ablation_study}. Note that the results were plotted by varying the classification behavioural threshold, which is the minimal amount of trajectory samples of a complete trajectory of an object that needs to be detected as abnormal in order for that complete trajectory to be labelled abnormal.

Results show that adding the adversarial learning framework on the pretrained DAE significantly improves the ADA as shown by the AUC. Also, adding additional features (height, width and orientation) into the data structure proves to better classify abnormal events from the normal ones. In addition, we observe that adding ALREC or additional features to the trajectory data does not affect the NDA.

\section{Conclusion}
In this article, we propose an adversarially learned reconstruction error classifier (ALREC) for detecting abnormal trajectory-based events by using a pretrained deep autoencoder (DAE) model. The adversarial method transforms a one-class DAE, which requires to manually define a detection threshold, to a two-class model that can classify normal and abnormal trajectories without that threshold. Compared to a typical GAN which is a generative model, our ALREC framework is a discriminative adversarial model that, instead of the generator generating fake trajectories, it rather generates abnormal MSEs that the discriminator learns to differentiate from the normal ones. An adversarial training algorithm is also proposed which makes the discriminator learn the most realistic looking abnormal MSEs while ensuring to learn the normal ones. The ALREC model was tested against other methods and was shown to outperform them. By conducting the ablation study, results show that all the components of our method ensure its success.


\section*{Acknowledgment}
This research was supported by a grant from IVADO funding program for fundamental research projects.



\bibliographystyle{IEEEtran}
\bibliography{IEEEabrv,bib/paper}

\end{document}